\documentclass{article}

\usepackage{PRIMEarxiv}

\usepackage[utf8]{inputenc} 
\usepackage[T1]{fontenc}    
\usepackage{hyperref}       
\usepackage{url}            
\usepackage{booktabs}       
\usepackage{amsfonts}       
\usepackage{nicefrac}       
\usepackage{microtype}      
\usepackage{lipsum}
\usepackage{fancyhdr}       
\usepackage{graphicx}       
\graphicspath{{media/}}     
\usepackage{algorithm}
\usepackage[noend]{algpseudocode}
\usepackage{amsmath} 
\usepackage{amssymb}

\usepackage{CJKutf8}

\title{Online Sequence Clustering Algorithm for Video Trajectory Analysis
\thanks{\textit{\underline{Citation}}: 
\textbf{Aximu Yuemaier, Xiaogang Chen, Xingyu Qian, Longfei Liang, Shunfeng Li and Zhitang Song. Online Sequence Clustering Algorithm for Video Trajectory Analysis. Pages.... DOI:000000/11111.}} 
}

\author{Aximu Yuemaier, Xiaogang Chen, Xingyu Qian, Longfei Liang,Shunfeng Li, Zhitang Song}
 
\begin{document}
\begin{CJK}{UTF8}{gbsn}
\maketitle

\begin{abstract}
Target tracking and trajectory modeling have important applications in surveillance video analysis and have received great attention in the fields of road safety and community security. In this work, we propose a lightweight real-time video analysis scheme that uses a model learned from motion patterns to monitor the behavior of objects, which can be used for applications such as real-time representation and prediction. The proposed sequence clustering algorithm based on discrete sequences makes the system have continuous online learning ability. The intrinsic repeatability of the target object trajectory is used to automatically construct the behavioral model in the three processes of feature extraction, cluster learning, and model application. In addition to the discretization of trajectory features and simple model applications, this paper focuses on online clustering algorithms and their incremental learning processes. Finally, through the learning of the trajectory model of the actual surveillance video image, the feasibility of the algorithm is verified. And the characteristics and performance of the clustering algorithm are discussed in the analysis. This scheme has real-time online learning and processing of motion models while avoiding a large number of arithmetic operations, which is more in line with the application scenarios of front-end intelligent perception.
\end{abstract}


\section{Introduction}
Behavioral perception of moving objects is a basic problem in computer vision applications. Among them, the modeling and prediction of trajectories are the focus of video perception and analysis\cite{yinwen01}. Especially in route prediction, predicting the future trajectory of pedestrians or vehicles in the next few seconds has received a lot of attention in road safety \cite{yinwen02}. This feature is also a key component in a variety of applications, such as autopilot\cite{yinwen03}, long-term object tracking\cite{yinwen04}, security monitoring\cite{yinwen05}, and robot planning\cite{yinwen06}. 

In video surveillance applications, tracking the condition of moving vehicles and pedestrians is very common, and large-track datasets can be created by storing the position of the observed objects over time. Data analysis in such datasets involves identifying and finding objects that move similarly or follow a particular pattern of motion. Clustering similar trajectories to produce representative examples can be an effective way to track and predict vehicle and human mobility\cite{yinwen07}. 

In the clustering learning scheme, the selection of similarity measure type has a great influence on the quality of clustering learning\cite{yinwen08}. For example, the longest common subsequence (LCS) measure is used in \cite{yinwen09} to cluster vehicle trajectories. There is also a way for \cite{yinwen10}\cite{yinwen11} to edit distances closely related to the LCS method. Another way to measure distance is dynamic time warping (DTW), which is validated in \cite{yinwen12}. LCS-based metrics are more robust to noise and outliers than DTW.

In the traditional trajectory-based learning behavior scenarios, there are two phases: observation/training and online evaluation \cite{yinwen13}. In the training phase, trajectory data needs to be extracted from a large number of videos to form a trajectory database, and then learn to establish patterns of similar movement behavior through clustering. Real-time motion behavior analysis is conducted during the evaluation phase using the models learned during the training phase. The size and completeness of the training stage track database in this scheme will affect the quality of clustering learning. In the real world, motion behavior changes over time, making it impossible for trained models to analyze and judge new untrained motion behavior. In addition, migrating such systems to new application environments requires new datasets and learning processes. For these clustered learnings based on trajectory datasets, the database management and learning algorithms need a lot of storage and computing costs\cite{yinwen07}. In an application environment, it is a big challenge to implement a real-time video analysis system with continuous learning ability.

In this work, a motion behavior perception algorithm based on online clustering of discrete sequences is proposed in the field of visual perception. The feature representation of the trajectory and data flow in the algorithm are discrete, which not only facilitates the design of the algorithm but also helps to improve the computational and storage efficiency. The learning of behavior patterns in this algorithm is incremental. The system can continuously learn new behavior patterns from new samples and can save most of the patterns that have been learned before. Moreover, the algorithm has the ability to forget infrequent patterns and remove noise data. 

\section{Algorithm Overview}
The main process of the visual trajectory-based behavior analysis system can be divided into three stages, as shown in Figure 1. This includes the generation of discrete trajectory sequences, model learning of online clustering, and model-based motion analysis. First stages, a video frame is extracted at a certain sampling rate, the moving target is detected from the image, and the discrete encoding is carried out in a certain way of segmenting the image. At the same time, target objects are tracked in real-time, resulting in discrete sequences of target objects in different areas. In the second stage, the discrete sequence generated in each region will be clustered by the first-layer nodes. The model number generated by the first layer node matching is combined with the image area number to generate a new discrete sequence to be sent to the second layer node. The final node uses coarser granularity to cluster the motion behavior trajectories in the picture for pattern learning.

\begin{figure}
\centering
\includegraphics[width=10.5 cm]{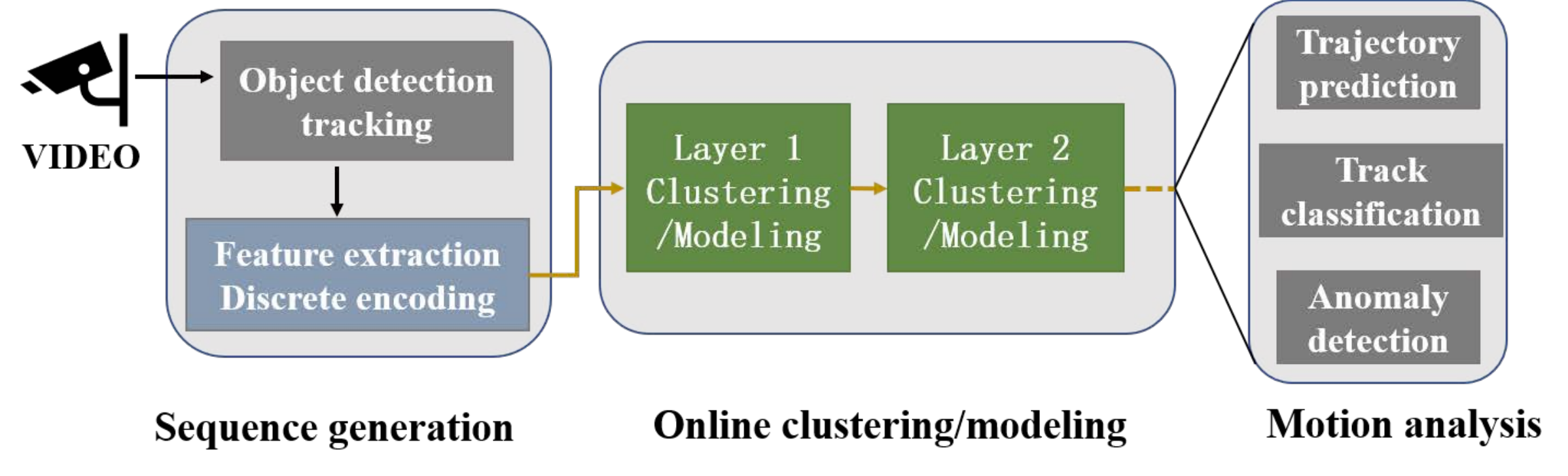}
\caption{Motion behavior analysis framework based on trajectory sequence clustering.\label{fig1}}
\end{figure}   

This paper mainly introduces the core part of the system algorithm, that is, around the online clustering algorithm and trajectory modeling. First, we introduce the discrete encoding method of image tracing and the composition of discrete sequences. Secondly, it focuses on the two-layer structure of the online clustering learning framework, such as similarity measurement, cluster-based modeling, and online model updating. Finally, the effectiveness of the algorithm is verified by real-time clustering learning of motion trajectories in the video environment.

\section{Online Cluster Learning Algorithm}
\subsection{Multi-layer structure of online cluster learning}
In the traditional trajectory-based motion behavior analysis model, the trajectory of each target needs to be completely preserved. However, for tasks such as behavior prediction, all the tracking points accumulated up to time t are not required, but only a small portion of the data window is used \cite{yinwen14}. In this work, we use image segmentation to focus only on the trajectories in this small area and use the discrete sequence generated by it to match and predict this motion model. Another advantage of using this split encoding method is that different coarse-grained classifications can be made by changing the partitioned encoding area, depending on the complexity of the environmental motion trajectories. Let's take the two-layer structure in Figure 2 as an example. The coding of a 4x4 splitter unit is processed at the first layer node, and the sequence stream sent at the first layer is processed as a 1x1 unit at the second layer node. This multi-layer processing not only alleviates the clustering learning tasks in each layer of nodes but also dynamically adjusts systems to meet the learning needs of different complex scenarios.

\begin{figure}[H]
\centering
\includegraphics[width=10.5 cm]{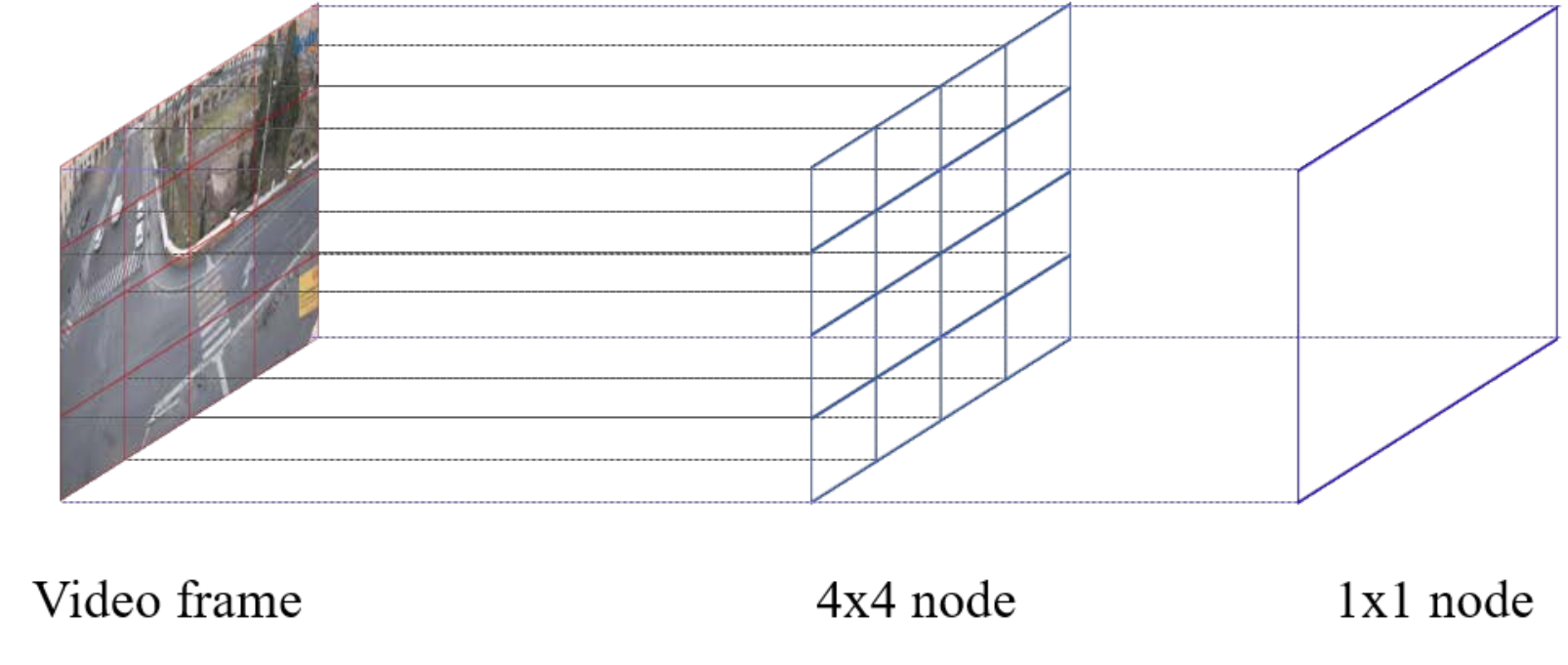}
\caption{Online clustering learning of two-layer processing.\label{fig2}}
\end{figure} 

\subsection{Discrete Sequence Generation}
In this processing system, the data extracted by the trajectory features and the data in the clustering process are propagated as discrete string sequences. Although the encoding and expression characteristics of discrete sequences are different in several stages of the system, they are clustered indiscriminately according to certain rules in the our algorithm. For example, in the first layer of the clustering module, the input sequence stream is generated after the feature extraction and encoding of the video image trajectory, and the processing result of this node is re-encoded to form the output sequence. 

The input sequence of the first layer node is generated by the discrete encoding of the trajectory of the motion behavior in the original video image. First, the video image is divided into several viewing fields by selecting a certain segmentation method, and then the motion trajectories are encoded in each area. Each split cell is encoded by four digits of the field as shown in Figure 3. The shaded part of the graph is the position of the moving object at the sampling time. Each partition unit can represent 16 states. Each split unit tracks a moving object, resulting in a discrete-time series that can represent the trajectory of motion within the split unit. The string sequence "0237540" produced by the trajectory encoding of a car passing through this field of view is shown here.

\begin{figure}[H]
\centering
\includegraphics[scale=0.3]{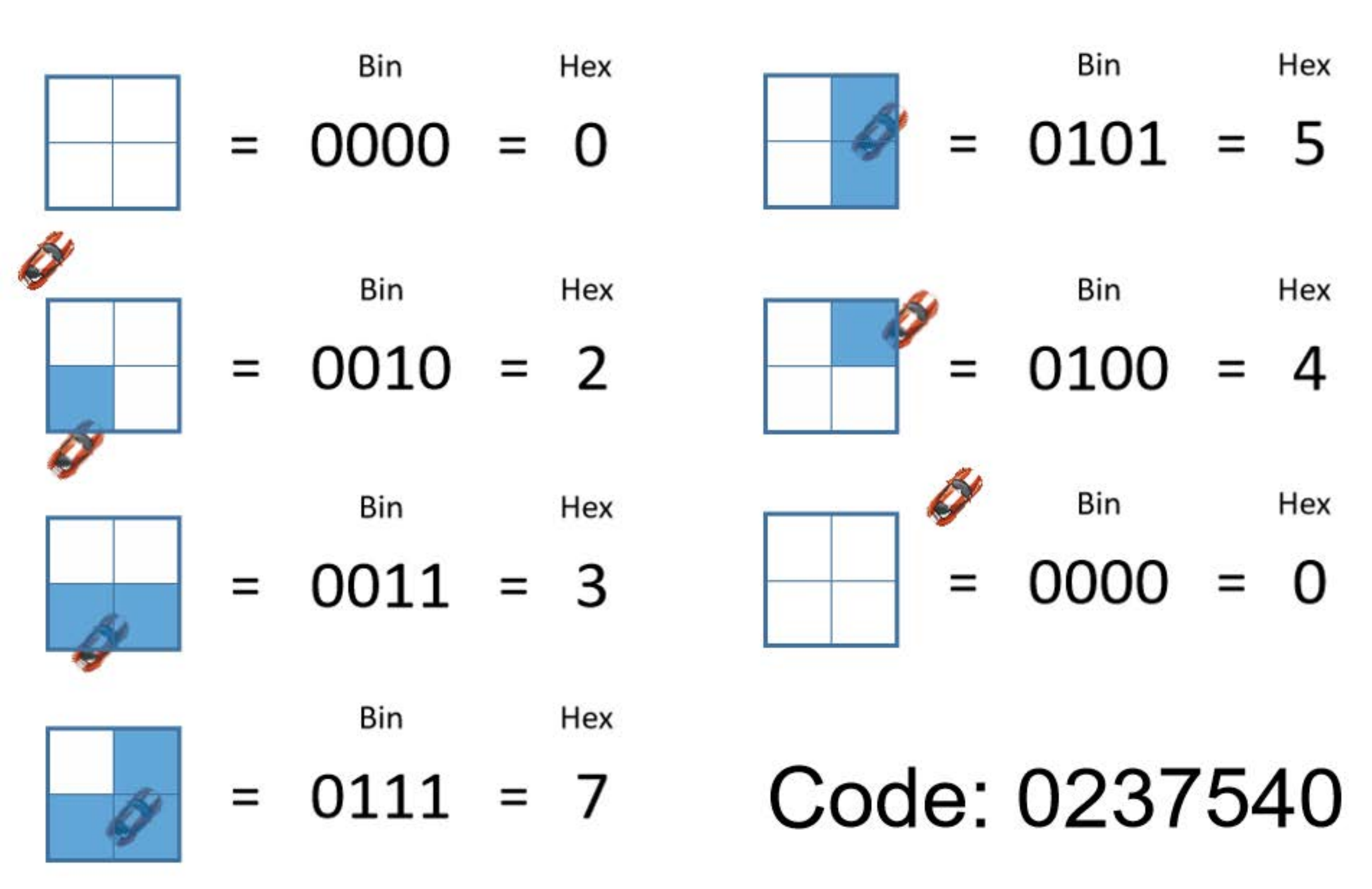}
\caption{An encoded sequence of images of car trajectories.\label{fig3}}
\end{figure}  

The encoding of the node output sequence, it is formed by merging two parts. One is the sequence model number that represents the current input sequence in the model library in this node, and the other is the information that represents the specific node number. For example, the output code of the node at the current time is S$_{o}$. The node number is N$_{i}$. The upper limit of the number of cluster models in a node is M$_{u}$. The model matched by the current input sequence is M$_{f}$. Then the output sequence can be encoded according to the following equation:

  \begin{equation}
  S_{o} = N_{i} \bullet M_{u} + M_{f}
  \end{equation}
  
Figure 4 shows the data flow input and output process of the two-layer structure. In the first layer, only two nodes are shown in the column. The input sequences S$^{'}$1$_{t}$ and S$^{'}$2$_{t}$ of these two nodes are encoded according to Figure 3. Inside the node, the clustering model similar to the input sequence of this node is found through sequence similarity matching. Then, the matching model number M$_{f}$ and the current node number N$_{i}$ are re-encoded according to Equation (1). S$^{'}$1$_{o}$ and S$^{'}$2$_{o}$ are the encoded data output by nodes 1 and 2 in the first layer. It should be explained here that only when the current sequence match changes in a node, the data sent to the next layer will be recoded and transmitted to the next layer, otherwise the output will not occur. That is, the current node remains inactive when the input sequence does not change. When there is a change in the input sequence, clustering and matching work begins inside the node. When second layer nodes receive data, these data are entered in the order of the previous layer node number. In the figure 4, \textit{Transfer} is used to represent the data transfer module, and the data of the previous node is entered to the next node in the order of the node number. The data flow direction inside the node of the second layer is basically the same as explained above. In general, through this discrete coding method, the coordination of nodes in each layer is more concise, and it also brings convenience to the processing of clustering algorithms within nodes.

\begin{figure}[H]
\centering
\includegraphics[width=10.5 cm]{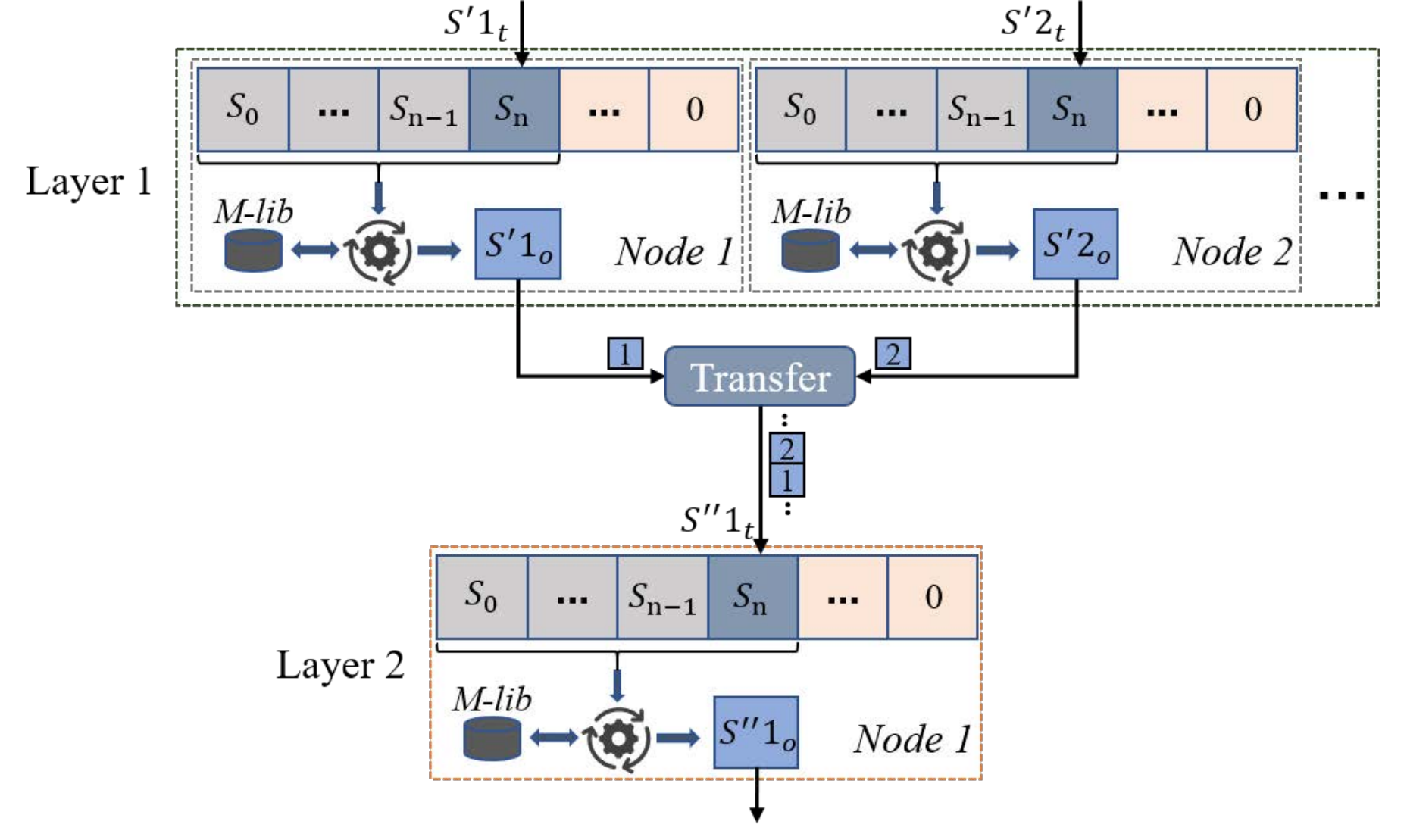}
\caption{Discrete sequence data flow with two-layer processing structure.\label{fig4}}
\end{figure}

\subsection{Online Sequence Streams Clustering}
The sequence stream clustering algorithm (seqStream) consists of two main phases. First, the similarity matching module matches the input sequence with the behavior model library in real-time and records the input behavior sequence. Secondly, when a set of sequences ends, the sequences are clustered, and new models are created or merged into corresponding model sequences. And periodically delete outdated models or merge similar models. In addition, noise data in the model sequence is periodically deleted to avoid conceptual drift in the behavior model sequence.
\\\textit{SeqStream} represents its micro-clusters (\textit{mc}) as 4-tuples:

 \begin{equation}
  mc~ = ~(t,~w,~SE,~SW)
  \end{equation}

Where \textit{mc.t} describes the last time the \textit{mc} was updated. \textit{mc.w} is used as a weight value to indicate how often new input sequences are merged into the cluster center (indirectly describing the importance of the cluster). Each time a new observation sequence is merged into an existing cluster, the corresponding weight is increased by 1. \textit{mc.SE} represents the discrete sequence stored in the model (micro-cluster). \textit{mc.SW} stores the weight of each character in \textit{mc.SE}, describing how often the corresponding characters are merged into the cluster (indirectly describing the importance of the characters in the cluster sequence). To account for variations in the distribution (conceptual drift), cluster weights decay exponentially each time a new observation occurs in the data stream: 

 \begin{equation} 
  mc.w = mc.w \bullet 2^{- \lambda(t_{now} - mc.t)}
  \end{equation}

The parameter $\lambda$ represents the fading (decay) factor, \textit{t$_{now}$} the recent time stamp, and \textit{mc.t} the time of the last micro-cluster update. \textit{t$_{gap}$} specifies the interval (number of new observations) and then triggers the cleanup process and micro-cluster merging. This cleanup procedure removes all \textit{mc} below a predefined weight threshold (i.e., clusters that were not recently updated) from the clustering result. At the same time, the noise data in the model sequence will be deleted according to the weight of the model sequence \textit{mc.WE}. Generally, the weights of noise data are quite different from those of normal data, so character data with large weight differences is deleted from the model sequence as noise data. This avoids the chaotic growth of the model sequence.

The clustering algorithm uses the distance measurement formula to measure the similarity between the behavior sequence and the existing behavior model. LCS is an important method to measure the similarity of time series\cite{yinwen15}. We use LCS to measure the similarity between the input sequence and \textit{mc} or between \textit{mc} and \textit{mc}. 

Suppose there are two string sequences S$_{x}$ and S$_{y}$, the length of which is n and m respectively, and the distance D(S$_{x}$,S$_{y}$) is given by the following equation:

 \begin{equation}
D\left( {S_{X},S_{Y}} \right) = 1 - sim\left( {S_{X},S_{Y}} \right) = 1 - \frac{LCS\left( {S_{X},S_{Y}} \right)}{\left. {{\min(}n,m} \right)}
  \end{equation}

LCS(S$_{x}$,S$_{y}$) is the longest common sub-sequence length of two sequences. The calculation method is as follows:

 \begin{equation}
  Lcs\left( {i,j} \right) =\! \left\{ \begin{matrix}
{\!0~~~~~~~~~~~~~~~~~;i = 0~or~j \!= \!0} \\
{~~~~\!\!\!\!norm\left( {mc.SW\lbrack i\rbrack} \right) \!+ Lcs\left( {i \!-\! 1,j \!-\! 1} \right)} ~~~~~~~~~~~~~~~~~~\\
{ ~~;x_{i} \!=\! y_{j},~i \!\geq \!1~or~j \!\geq \!1} \\
{max\left\{ \begin{matrix}
\left. {Lcs\left( i \!-\! 1,j \right.} \right) \\
\left. {Lcs\left( i,j \!-\! 1 \right.} \right) \\
\end{matrix} \!\right.} ~~~~~~~~~~~~~~~~~~~~~~~~~~~~~~~~~~~~~~~\\
{~~;x_{i} \!\neq\! y_{j},~i \!\geq\! 1~or~j \!\geq \! 1}
\end{matrix} \right.
\end{equation}

Because the generation of each sequence in the model is incremental, even though the characters in the two sequences are similar, due to the different weights of the two characters, the two characters may not be able to add to the number of similar characters for the similarity of the whole sequence. To reduce the influence of noise sequences on similarity matching and eliminate the influence caused by different decay speeds of different model sequence weights, LCS calculations are performed by adding the normalized values of the weights of the matching features.

When the newly observed input sequence needs to match the model or needs to create a new model, both \textit{mc.w} and \textit{mc.WE} can be set to 1. The closed \textit{mc} is selected to be merged when its distance falls below a predefined threshold $\epsilon$. In the merger process, \textit{mc.t} is updated to the current time t, and \textit{mc.w} is added by 1. \textit{mc.SE} is formed by merging two sequences. The positions of similar characters do not need to be moved, and the different characters are inserted into the character sequence in order. The weights of similar characters in \textit{mc.WE} are added to each other, and the weights of different characters are inserted into the weight sequence in order.

\begin{algorithm}[!ht]
    \caption{seqStream.}
    \hspace*{0.02in} {\bf Require:}
    $ t_{gap}, \epsilon,  \lambda, \mu $ \\
    \hspace*{0.02in} {\bf Initialize:}   
    $S=\varnothing, MC=\varnothing$
    \begin{algorithmic}[1]

    \While{ the stream is active} %
　　      \State $t_{now} \leftarrow $ current timestamp
         \State Read new sequence $S_{t}$ from the stream at time $t_{now}$
         \If{$(|S|=0)$ or $(S_{t}!=S_{n-1}) $}  
　　　　        \If{$S_{t}$ != 0 }  
　　　　            \State Add $S_{t}$ to the sequence list $S$
                   \If{$|MC| > 2$} 
                        \For{each micro-cluster $mc \in ~ MC$}
                              \State  $dist \leftarrow D(S, mc)$
                        \EndFor
                        \State $mc_{cl}  \leftarrow $ closest $mc$ according to dist
                        \State  $S_{p} , mode_{num} \leftarrow $predict according to $mc_{cl}$
                   \EndIf
　　          \Else
                    \If{$|S| != 0$} 
                        \State $\left. {mc}_{new}\leftarrow\left( t_{now}~,1~,~~S~,~~{SW}_{init}~ \right) \right.$
                        \If{$|MC| > 2$} 
                            \If{$D(mc_{new}, mc_{cl}) \leq \epsilon $} 
                                \State $mc_{cl} \leftarrow $ MERGE($mc_{new}, mc_{cl}$)
                             \Else
                                \State  Add $mc_{new}$ to set of cluster $MC$
                            \EndIf
                        \Else
                            \State Add $mc_{new}$ to set of cluster $MC$
                        \EndIf
                        \State Initialize $S=\varnothing$
                    \EndIf
　　　　            
　　          \EndIf
         \EndIf
         
　　      \If{$t_{now}$ mod $t_{gap} = 0$}
　　　　      \State  CLEANUP(.)
　　      \EndIf
    \EndWhile
    \end{algorithmic}
\end{algorithm}

\subsection{Algorithm Specification}
The pseudocode of our online clustering algorithm is shown in Algorithm 1. First, the algorithm reads the character at the current time from the data stream (line 3). When processing sequence, first judge whether the current character is ‘0’ (because the behavior sequence starts and ends with the character ‘0’). If the received character is not ‘0’, the sequence of the input part will enter the model matching process(lines 6-10), and the matching model number will be output according to the demand, and the next character will be predicted. If the received character is 0, the process of creating or merging a new model will be executed (lines 12-22), where the merging process of the model is incremental. Finally, after the time interval of t$_{gap}$, clean and merge the model library.

The procedure for cleaning up the model library is shown in algorithm 2. First is the model cleaning process (lines 2-5). The \textit{mc.w} of all models in the model library is attenuated according to equation (5), and then the threshold value of line 4 is used to determine whether to clean from the model library. After that, the cleaning process of noise characters in the model (lines 7-10), according to the large difference between the weight of noise characters and other characters, if the difference is greater than the given threshold $\mu$, cleans the corresponding characters and their weights carefully.
Finally, it is necessary to judge whether there is a similarity between the models and merge the corresponding similar models.
 
\begin{algorithm}[!tb]
    \caption{Cleanup step.}
    \begin{algorithmic}[1]
    \Function{CLEANUP}{$.$}  
        \For{each $mc \in MC$}
            \State $mc.w \leftarrow mc.w \bullet ~2^{- \lambda(t_{now} - mc.t)}$
             \If {$mc.w \leq ~2^{- \lambda t_{gap}}$)}
                   \State  remove $mc$ from $MC$
             \Else
                    \State equal ratio scales every weight in $mc.WE$
                        \For{each $mc.we[i]$}
                            \If {MAX$(mc.WE) - mc.WE[i] > \mu $}
                                \State remove $mc.SE[i]$ and $mc.WE[i]$ from $mc$
                            \EndIf
                        \EndFor
             \EndIf
        \EndFor 
        \State Merge all ${mc}_{i}$, ${mc}_{j}$ where $D({mc}_{i}, {mc}_{j}) \leq \epsilon $
  
    \EndFunction
    \end{algorithmic}
\end{algorithm}

The final algorithm 3 is the pseudocode of the model merge process. First, the longest common sub-sequence is calculated according to the dynamic programming algorithm (line 3), and the LCS calculation output table (lcs\_dp) is cached. Then use the recursive algorithm \textit{LCS\_GEN (*)} to find the position of similar characters (algorithm4), and insert and merge the characters in the two sequences into mc$_{merge}$ in order.

\begin{algorithm}[!tb]
    \caption{Merge step (1).}
    \begin{algorithmic}[1]
    \Function{MERGE}{${mc}_{1},{mc}_{2}$}  
    \State ${mc}_{merge} \leftarrow(0~,0~,~\varnothing~,~\varnothing~)$   
    \State $Lcs\_dp \leftarrow LCS\_TAB({mc}_{1}.SE~,{mc}_{2}.SE)$  
    \State ${mc}_{merge} \leftarrow LCS\_GEN(Lcs\_dp, {mc}_{1}, {mc}_{2},  \newline 
    ~~~~~~~~~~~~~~~~~~~~~~~~~~~~~~ |{mc}_{1}.SE|, | {mc}_{2}.SE|, {mc}_{merge})  $ 
    \State ${mc}_{merge}.t \leftarrow{mc}_{1}.t $
    \State ${mc}_{merge}.w \leftarrow{mc}_{1}.w + 1  $
    \State \Return ${mc}_{merge}$
    
    \EndFunction
    \end{algorithmic}
\end{algorithm}

\begin{algorithm}[tb]
    \caption{Merge step (2).}
    \begin{algorithmic}[1]
    \Function{LCS\_GEN}{$Lcs\_dp, {mc}_{1}, {mc}_{2},   
     i, j, {mc}_{merge}$}  
    \If {$i==0$}
    \For{$k$ in $range(j)$}
            \State add ${mc}_{2}.SE[k]$ to ${mc}_{merge}.SE$
            \State add ${mc}_{2}.WE[k]$ to ${mc}_{merge}.WE$
        \EndFor     
     \State \Return
    \ElsIf{$j==0$}
    \For{$k$ in $range(i)$} 
         \State add ${mc}_{1}.SE[k]$ to ${mc}_{merge}.SE$ 
         \State add ${mc}_{1}.WE[k]$ to ${mc}_{merge}.WE$
    \EndFor
     \State \Return
    \EndIf
    \If {$lcs\_dp[j][i]==lcs\_dp[j-1][i]$} 
      \State $LCS\_GEN(Lcs\_dp, {mc}_{1}, {mc}_{2}, i, j-1, {mc}_{merge})$
      \State add ${mc}_{2}.SE[j-1]$ to ${mc}_{merge}.SE$
      \State add ${mc}_{2}.WE[j-1]$ to ${mc}_{merge}.WE$
    \ElsIf {$lcs\_dp[j][i]==lcs\_dp[j][i-1]$}
      \State $LCS\_GEN(Lcs\_dp, {mc}_{1}, {mc}_{2}, i-1, j, {mc}_{merge})$
      \State add ${mc}_{1}.SE[i-1]$ to ${mc}_{merge}.SE$
      \State add ${mc}_{1}.WE[i-1]$ to ${mc}_{merge}.WE$
    \Else
      \State $LCS\_GEN(Lcs\_dp, {mc}_{1}, {mc}_{2}, i-1, j-1, {mc}_{merge})$
      \State add ${mc}_{1}.SE[i-1]$ to ${mc}_{merge}.SE$
      \State add ${mc}_{1}.WE[i-1] + {mc}_{2}.WE[j-1]$ to ${mc}_{merge}.WE$
    \EndIf

    \EndFunction
    \end{algorithmic}
\end{algorithm}

\section{Evaluation}
\subsection{Experimental setup}
To evaluate our algorithm, we built a clustering algorithm verification environment with python. All experiments are performed on an Ubuntu machine with an Intel i7-10700 CPU with 2.90GHz. For the analysis, we utilize real-world data sets. The video data collected by the webcam \cite{yinwen16} is the experimental data set. The video image is the monitoring image of the intersection as shown in Figure 5, which can regularly display the movement behavior model of vehicles and pedestrians.

\begin{figure}[H]
\centering
\includegraphics[scale=1]{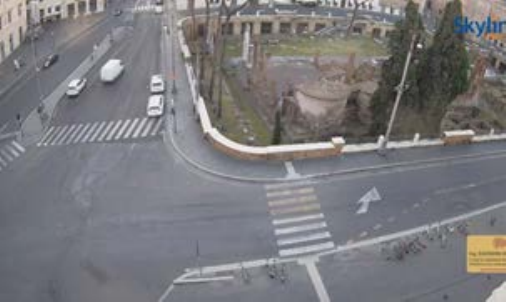}
\caption{Video image of intersection monitoring.\label{fig5}}
\end{figure}

In the clustering algorithm seqStream, four hyperparameters need to be determined in advance. They are similar threshold parameter $\epsilon$, attenuation rate (forgetting rate) $\lambda$, threshold $\mu$ of noise character weight judgment, and clearing time interval t$_{gap}$.

The clustering quality of the algorithm is measured by the correct clustering rate (CCR) \cite{yinwen17}:

 \begin{equation}
CCR = \frac{1}{N}{\sum\limits_{c = 1}^{k}p_{c}}
  \end{equation}

Where N represents the total number of trajectory sequences involved in all experiments, p$_{c}$ represents the total number of behavior trajectories matching the c-th cluster.

\subsection{Experimental Results}
First, the impact of the similarity metric threshold $\epsilon$ on the clustering results of the model is examined. The experimental video picture is divided into 64 nodes in the first layer, and the 3647 discrete sequences generated by the resulting encoding are used as samples for clustering matching experiments. In the experiment, $\lambda$=1e$^{-2}$, $\mu$=10, and t$_{gap}$=20 were specified.

Figure 6 shows the clustering quality of the behavior sequence when taking different $\epsilon$ values. It can be seen that when the $\epsilon$ value is between 0.2 and 0.5, it has a higher CCR and a lower variance. When the $\epsilon$ value is too small, the algorithm learns too many behavioral models. Not only does this consume too many computational storage resources, but noisy data can affect the results of model matching. When the $\epsilon$ value is too large, the resolution of the behavior trajectory decreases, and the discrimination of the behavior model decreases. The threshold value of the subsequent experiment in this article is set to 0.3 for $\epsilon$.

\begin{figure}[H]
\centering
\includegraphics[width=10.5 cm]{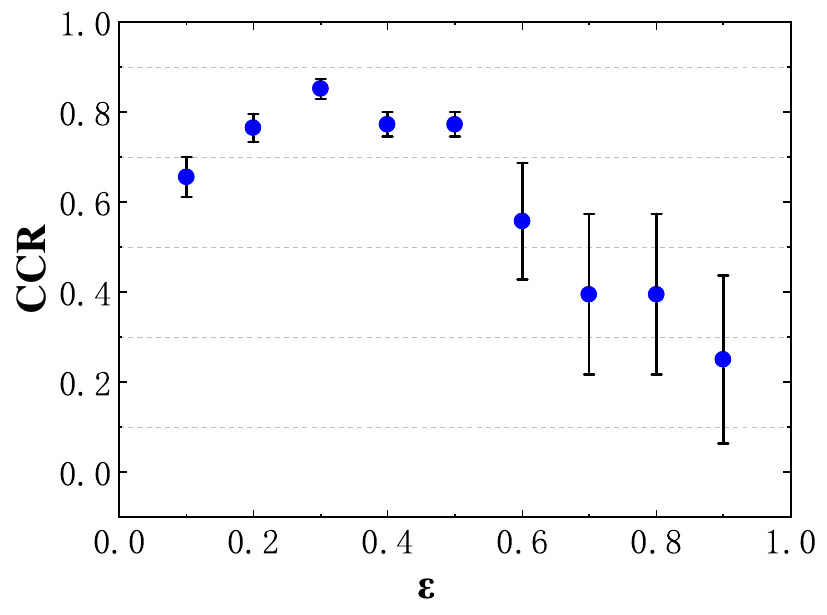}
\caption{The clustering quality of the behavior sequence when different $\epsilon$ values; $\epsilon$ values between 0.2 and 0.5 have higher CCR and lower variance.\label{fig6}}
\end{figure}   

Next, the seqStream algorithm compares the performance and clustering quality of the following two traditional clustering algorithms:
(1)	DBCSAN: DBCSAN\cite{yinwen18} is a representative density-based clustering algorithm. For string similarity, Levenshtein distance is used as a metric. And use scikit-learn's dbscan for string clustering.
(2)	Hierarchical clustering: Hierarchical clustering attempts to divide the sample data set on different ‘levels’, layer-by-layer clustering \cite{yinwen19}. The bottom-up condensed hierarchical clustering method is adopted, and the similarity between each cluster is measured by Levenshtein distance.

As shown in Table 1, the three perspectives of CCR, time, and required storage space are compared. Where N is the number of sample sequences and M is the number of established models. As can be seen from the table, our algorithm not only has high clustering quality but also has great advantages in running time and storage space.

\begin{table}
 \caption{Clustering algorithm comparison}
  \centering
  \begin{tabular}{llll}
    \toprule
    \textbf{Algorithm}	& \textbf{CCR(\%)}	& \textbf{Time(S)}	& \textbf{Memory}\\
    \midrule
DBCSCAN		& 81.2 & 0.9 & O(N)\\ 
Hierarchical		& 98.5 & 0.26 & O($N^2$)\\
Ours		& 92.8 & 0.18 & O(M)\\ 
    \bottomrule
  \end{tabular}
  \label{tab:table}
\end{table}

\subsection{Online Cluster Learning Process}
Our algorithm is an incremental learning process based on discrete sequence clustering, which dynamically changes the number of behavioral models and model sequences as the data flow passes. Due to the regular cleaning and merging process of models in the algorithm, it is not sensitive to noisy data and less sensitive to sequence input order. For example, the trajectory sequence '4CEA20' is generated when the vehicle passes through the red segmented area in Figure 7. In the fusion process based on clustering models, the weights of these characters will be high, and this sequence can be seen as the centroids within the cluster.

\begin{figure}[H]
\centering
\includegraphics[scale=1]{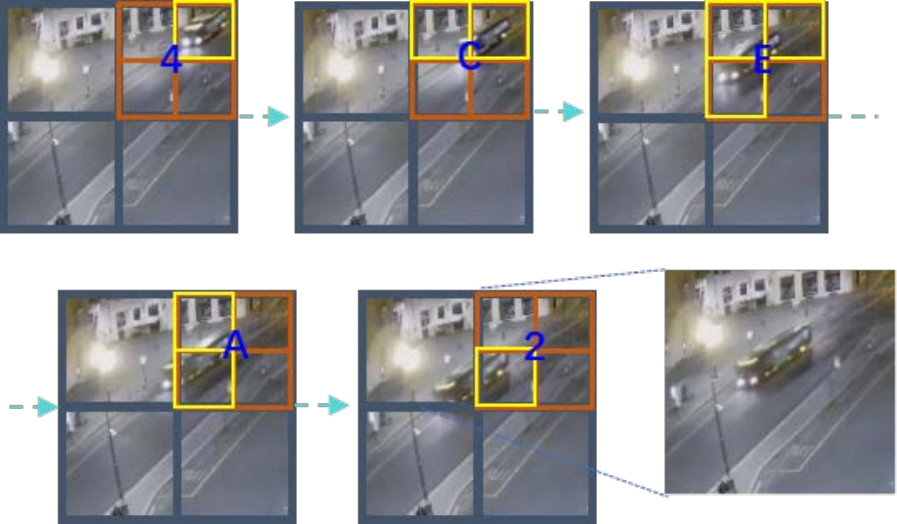}
\caption{Discrete sequences of motion behavior “4CEA20”.\label{fig8}}
\end{figure}

Figure 8 shows the process of model sequence merge and clear at different times for this sequence model. The color depth in different boxes in the figure represents weight value W$_{i}$. Due to the influence of noisy data, the sequence merged by the model sequence at different times is not the same. Due to the proximity of the vehicle's driving area to the street lights in Figure 7, there are certain difficulties in encoding the motion trajectory. For example, at time t$_{2}$, t$_{3}$, t$_{4}$, noise data such as' 8 'and' 5 'were added to the model sequence. However, with the clustering merge and clear process, these noisy data will be removed from the model sequence. It can be seen that the characters corresponding to the correct trajectory sequence always have a high W$_{i}$ value that can better match the corresponding behavior trajectory.

\begin{figure}[H]
\centering
\includegraphics[scale=0.8]{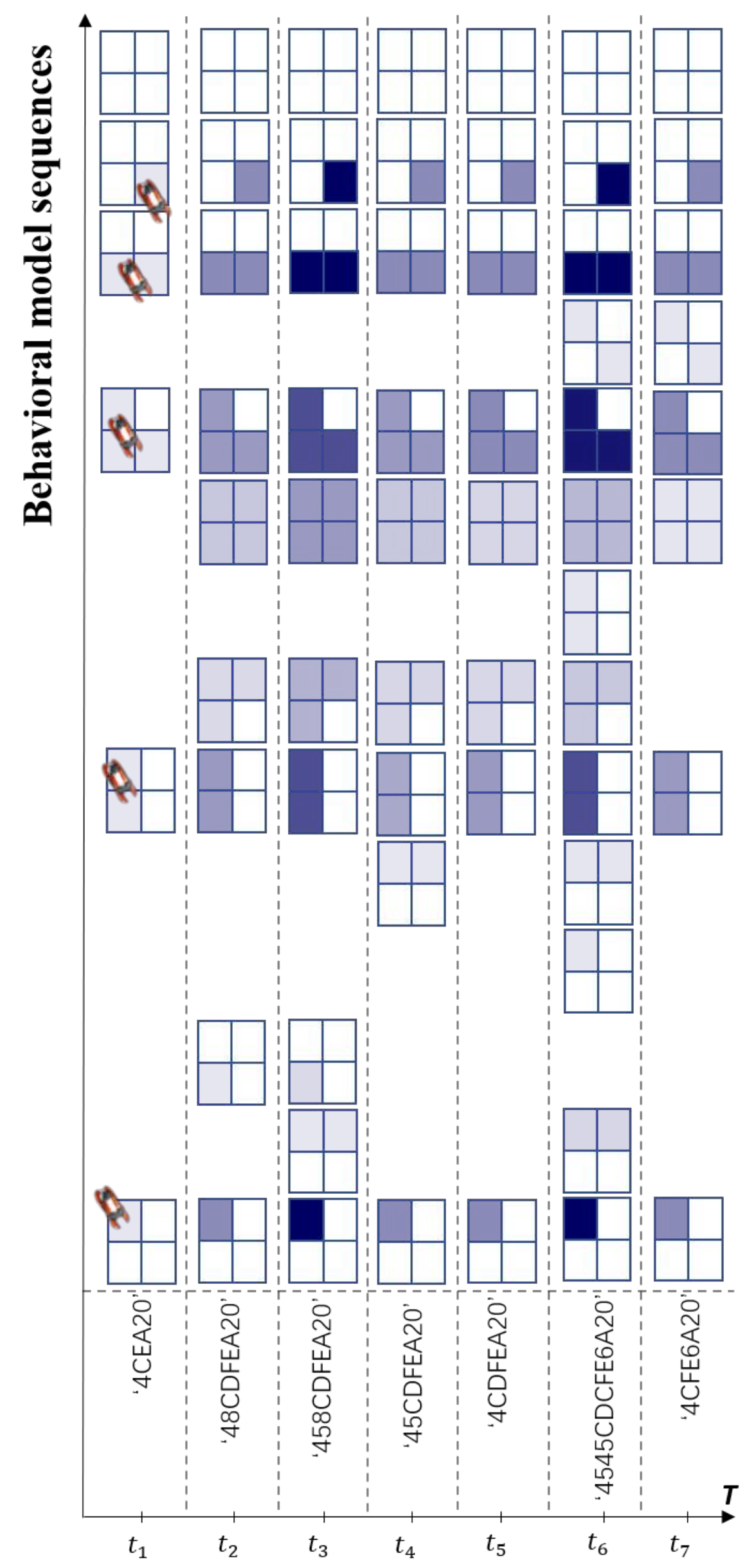}
\caption{The model matched by the sequence '4CEA20' and the online learning process of this model; the changes of the model sequence at 7 time points were selected respectively.\label{fig9}}
\end{figure}

We extracted 230 sets of motion behavior sequences from the motion sequence encoded data in the video as experimental data. In order to investigate the incremental learning process of the algorithm and the influence on the sequence input order and noise, the experiment of random input of the first-layer node sequence is carried out on the basis of these 230 sets of sequences. 
 
 The experimental results show that the clustering accuracy can finally be stable above 90\%. We selected four typical random input experimental results for display, as shown in Figure 9.
The horizontal ordinate is the number of motion sequence streams flowing in over time, and the ordinate is the clustering quality. As can be seen from the figure, the clustering learning accuracy is low at first, and the clustering quality tends to stabilize after 25 to 50 sequence inputs. This is due to reasons such as background lighting in the video, even if similar movement behavior, several characters appear different when encoding. Therefore, at the beginning of the system operation, similar sequences are judged to be different behavioral models under the threshold condition of low similarity distance. However, with the clustering learning of the algorithm, those characters that often appear in similar sequences are strengthened (the weight becomes larger), so these characters with larger weights occupy a larger proportion in similar matching, and the clustering accuracy will be correspondingly improved.
As shown in the C experimental process in the figure, the clustering quality drops to 0.4 at the beginning, that is, the system is greatly affected by the noise data, but as the input sequence stream increases, it eventually returns to a higher clustering accuracy rate. The results of the A and D experiments were similar, and they quickly returned to the optimal state, indicating that the distance between the sequence streams they began to input was scattered. For the group B experiment, although its clustering quality can quickly return to a stable state, its overall clustering quality is below 0.9. It starts to trend upward later, and if the system continues to learn online, the clustering quality may improve. 

To sum up, our algorithm has strong online learning capabilities and can continue to learn while not being sensitive to noise data and input order.
If an input trajectory sequence matches the wrong model at a certain moment, the predicted behavior will also be correspondingly incorrect. However, abnormal noise data does not last long, and abnormal sequences occupy a small portion of all learning sequences. Therefore, after learning for a period of time, the same sequence with incorrect judgment will have correct model matching and prediction.

\begin{figure}[H]
\centering
\includegraphics[width=10.5 cm]{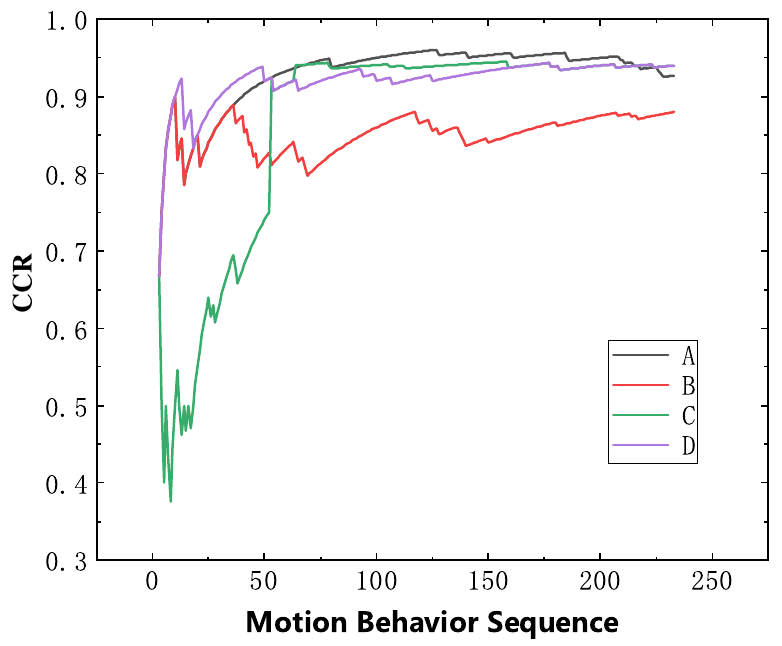}
\caption{The clustering quality of motion behavior sequence over time.\label{fig7}}
\end{figure}

\section{Discussion and Future Work}
In this work, we introduce an algorithm for online learning and real-time analysis of unsupervised visual scenes based on the repetitive structure inherent in learning motion trajectories. The representation of trajectory features and data flow in the algorithm is discretely encoded, which not only facilitates the design of the algorithm but also helps to improve the calculation and storage efficiency of the algorithm. The proposed online cluster learning algorithm enables the system to have a continuous learning ability and a certain adaptive ability to external noise. Future work will further study the adaptive adjustment of hyperparameters in the algorithm and verify the reliability and scalability of the system in the heterogeneous scene.

\bibliographystyle{unsrt}  
\bibliography{Arxiv}  

\end{CJK}
\end{document}